\documentclass[conference]{IEEEtran}

\usepackage{cite}
\usepackage{amsmath,amssymb,amsfonts}
\usepackage{algorithm}
\usepackage{algorithmic}
\usepackage{graphicx}
\usepackage{textcomp}
\usepackage{xcolor}
\usepackage{booktabs}
\usepackage{multirow}
\usepackage{adjustbox}
\usepackage{array}
\usepackage{url}
\usepackage{balance}
\usepackage{hyperref}
\usepackage{caption}
\usepackage{subcaption}
\usepackage{mathtools}
\usepackage{xcolor,colortbl}
\definecolor{ieeeblue}{RGB}{0,102,204}
\definecolor{lightgreen}{RGB}{0,150,0}
\definecolor{darkgreen}{RGB}{0,120,0}
\definecolor{llg}{RGB}{60,140,60}
\begin{document}

\title{QFCQT: A Chaotically Gated Quantformer Framework for Volatile Time-Series Forecasting}
% \author{Anonymous Author(s)}

\author{
\IEEEauthorblockN{Junkai Lin$^\dagger$, Siqi Hou$^\dagger$, Raymond Lee$^*$}
\IEEEauthorblockA{
Faculty of Science and Technology, Beijing Normal-Hong Kong Baptist University, Zhuhai, China\\
Guangdong Provincial Key Lab of Interdisciplinary Research and Application for Data Science (BNBU), Zhuhai, China\\
Email: raymondshtlee@bnbu.edu.cn
}
}

\maketitle

\begingroup
\footnotesize
\noindent
© 2026 IEEE. Personal use of this material is permitted.
Permission from IEEE must be obtained for all other uses,
including reprinting/republishing this material for advertising
or promotional purposes, creating new collective works,
resale or redistribution to servers or lists,
or reuse of any copyrighted component of this work in other works.

\vspace{0.5em}
\endgroup

\begin{abstract}
Forecasting non-stationary time series remains difficult due to long-range dependencies, local volatility bursts, structural shifts, and nonlinear oscillatory behaviors. Although Transformer-based forecasters are effective for modeling long-term temporal dependencies, their feed-forward blocks typically rely on smooth static activations that are insufficiently sensitive to abrupt regime changes. Motivated by quantitative Transformer designs and oscillator-based nonlinear activations, we propose QFCQT, short for Quantum-Fractal-inspired Chaotically Gated Quantformer, for robust forecasting under complex volatile dynamics. Here, “quantum-fractal-inspired” denotes a computational analogy based on soft oscillator superposition and multi-scale nonlinear responses, rather than a formal quantum-mechanical or fractal-theoretic derivation.

QFCQT consists of three main components: 1) a Quantformer-style numerical encoder that directly processes multivariate inputs via linear embedding; 2) a learnable Lee-oscillator activation module that maps scalar pre-activations to dynamic oscillatory responses and summarizes them through Max-over-Time pooling; and 3) a smooth-chaotic gated fusion mechanism that adaptively balances conventional smooth activations and chaos-sensitive responses. Furthermore, instead of using a single fixed oscillator, QFCQT employs a soft superposition of eight parameterized Lee oscillator families to adaptively capture different nonlinear response patterns across regimes.

Experiments on ETTh$_1$, ETTh$_2$, and A-share Stock Index benchmarks show that QFCQT consistently outperforms strong baselines, including Informer, LogTrans, LSTMa, HAT, and COTN. On highly volatile settings, the proposed method achieves up to 43.9\% MSE improvement over HAT and 41.3\% over COTN on ETTh$_2$ with prediction horizon 24. These results demonstrate that chaos-aware activation dynamics integrated into a Quantformer-style numerical backbone provide an effective and robust solution for forecasting in non-stationary and highly volatile environments.
\end{abstract}


\section{Introduction}
Time-series forecasting is an important problem in quantitative finance, power systems, traffic analysis, and industrial monitoring. In practice, time series often exhibit long-range dependency, local fluctuation, volatility clustering, abrupt structural changes, and non-stationary behavior, which make accurate prediction difficult.

Recurrent models such as LSTM and GRU can model sequential dependence~\cite{zeng2025dfc}, but their sequential computation limits parallelism and often makes long-horizon optimization difficult. Transformer-based models address this issue through self-attention and have become widely used in long-term forecasting. However, most existing Transformer forecasters mainly improve attention design or computational efficiency, while the nonlinear transformation in the feed-forward block remains largely unchanged. Under highly volatile conditions, such smooth static activations may be insufficient to capture abrupt local transitions.

Recent studies suggest two relevant directions. \textit{Quantformer} (Quantitative Transformer)~\cite{zhang2024quantformer} shows that numerical time series can be modeled more naturally by direct linear embedding instead of following the standard NLP-oriented Transformer pipeline. COTN~\cite{tang2025cotn} shows that Lee Oscillator~\cite{lee2004transient} dynamics, Max-over-Time pooling, and gated fusion can improve responsiveness to chaotic and highly volatile signals. Motivated by these observations, we propose \textbf{QFCQT}, a \textbf{Q}uantum-\textbf{F}ractal-inspired \textbf{C}haotically gated \textbf{Q}uantformer framework for quantitative \textbf{T}ime-series prediction.

The proposed method retains temporal self-attention for global dependency modeling, while replacing the conventional pointwise activation in the feed-forward block with a chaotically gated activation based on multiple Lee oscillator families. Specifically, QFCQT adopts linear embedding for multivariate numerical sequences, introduces a soft superposition of eight parameterized Lee oscillator families with Max-over-Time pooling, and uses smooth-chaotic gated fusion with surrogate-gradient training for stable optimization.

Our contribution is therefore positioned as a practical integration and extension of existing ideas rather than a new formal theory of quantum or fractal dynamics. Specifically, QFCQT combines a Quantformer-style numerical encoder with a learnable mixture of Lee-oscillator activation families and a smooth-chaotic gated fusion mechanism. This design aims to improve the sensitivity of Transformer feed-forward blocks to local volatility, abrupt regime changes, and nonlinear oscillatory patterns while preserving stable optimization.

\section{Related Work}

\subsection{Transformer-Based Time-Series Forecasting}

Transformer variants have been widely used in time-series forecasting because self-attention can effectively model long-term temporal dependencies. Existing methods improve forecasting performance through sparse attention, decomposition mechanisms, frequency-domain modeling, patch tokenization, or architectural efficiency. Despite these advances, many of them still employ conventional smooth activations inside the feed-forward network, which may reduce responsiveness when local system dynamics change rapidly.

\subsection{Quantformer for Numerical Sequences}
Quantformer demonstrates that standard NLP-style Transformer design is not always ideal for numerical time-series modeling. In particular, it replaces word embedding with a learnable linear projection for quantitative inputs and simplifies the output process by removing autoregressive decoding and masking operations, making the framework better suited to stock-related numerical sequences. 

This principle is directly relevant to forecasting tasks where the input tokens are already temporally ordered numerical observations rather than symbolic language units. Our work inherits this idea and extends it from stock-oriented factor learning to a broader multivariate forecasting setting.

\subsection{Chaotic Oscillator-Based Neural Activation}
Lee Oscillator-based neural dynamics provide a promising mechanism for modeling systems with strong nonlinearity, volatility, and abrupt transitions. COTN~\cite{tang2025cotn} shows that chaotic activation can be made practical in Transformer forecasting by compressing oscillator trajectories using Max-over-Time pooling and adaptively fusing the resulting response with GELU through a learnable gate.

Moreover, the Lee Oscillator is particularly attractive because its discrete-time formulation is computationally tractable and its progressive chaotic growth is suitable for volatile systems. 

Our method builds on this foundation, but instead of relying on a single chosen oscillator type, we introduce a learnable soft mixture over eight oscillator families and integrate it into a Quantformer-style backbone.

\section{Methodology}

\subsection{Problem Formulation}
Given a multivariate historical sequence
\begin{equation}
\mathbf{X} \in \mathbb{R}^{T \times C},
\end{equation}
where $T$ is the look-back window and $C$ is the number of input variables, the objective is to predict a future sequence
\begin{equation}
\mathbf{Y} \in \mathbb{R}^{H \times D},
\end{equation}
where $H$ is the forecasting horizon and $D$ is the number of target variables. The model learns a mapping
\begin{equation}
f_{\Theta}: \mathbb{R}^{T \times C} \rightarrow \mathbb{R}^{H \times D}.
\end{equation}

\subsection{Overall Framework}
\begin{figure*}[hbt!]
    \centering
    \includegraphics[width=0.7\textwidth]{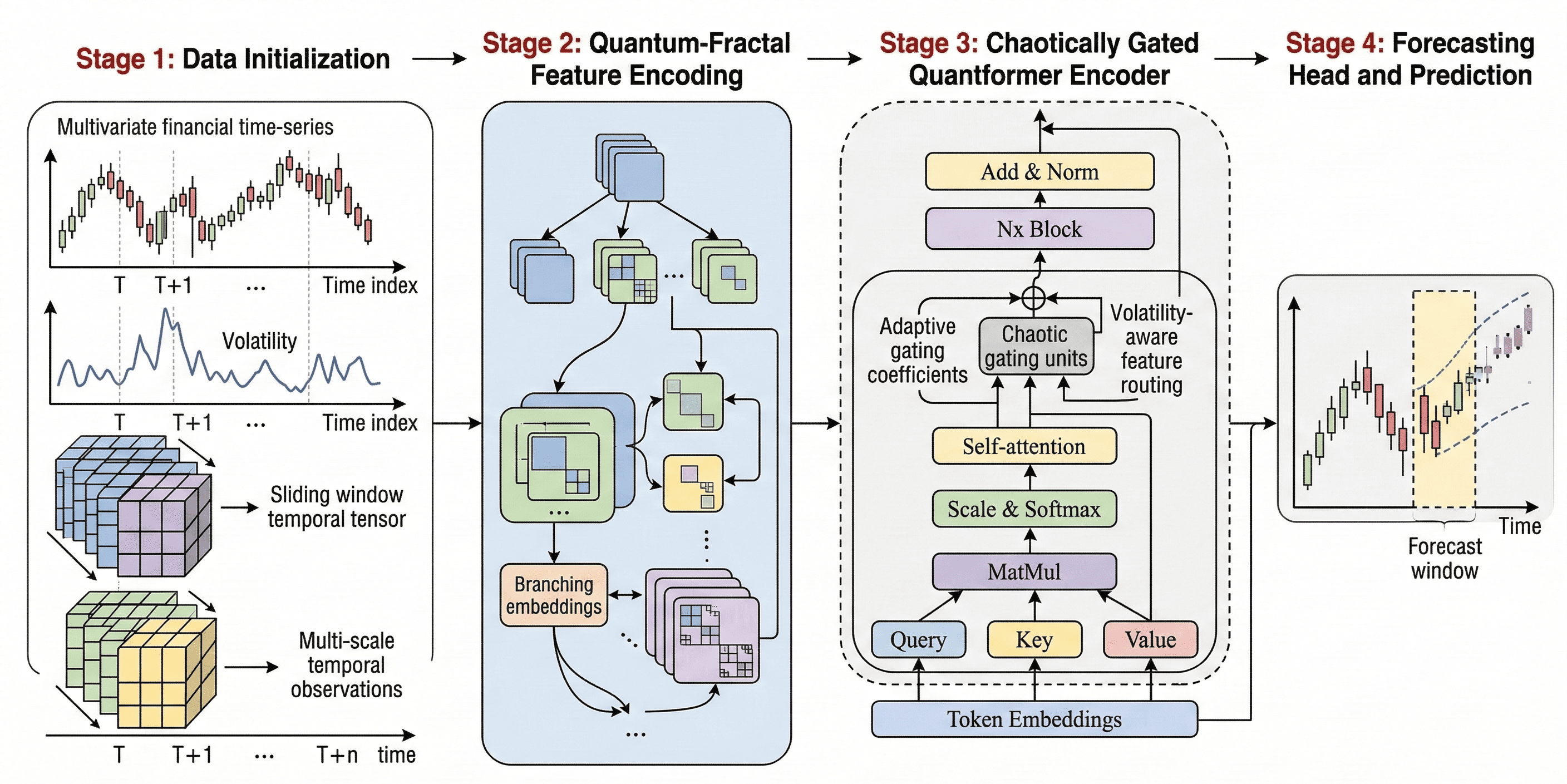}
\caption{Overview of the proposed \textbf{QFCQT} framework for time-series forecasting.}    
\label{fig:frame}
\end{figure*}

The proposed QFCQT framework consists of four stages, as illustrated in Fig.~\ref{fig:frame}, QFCQT follows a four-stage forecasting pipeline: numerical sequence initialization, quantum-fractal-inspired feature encoding, chaotically gated Quantformer encoding, and horizon projection. The feature encoding stage denotes multi-scale numerical representation and branching embedding organization, rather than a formal quantum operator or mathematically defined fractal transform. The chaotically gated encoder then combines temporal self-attention with oscillator-based gated activations to capture both long-range dependencies and local volatile transitions.

The hidden representation of the $\ell$-th block is written as
\begin{align}
\mathbf{H}^{(0)} &= \mathbf{X}\mathbf{W}_{e} + \mathbf{b}_{e}, \\
\widetilde{\mathbf{H}}^{(\ell)} &= \mathrm{AttnBlock}(\mathbf{H}^{(\ell-1)}), \\
\mathbf{H}^{(\ell)} &= \mathrm{QFCQT\text{-}FFN}(\widetilde{\mathbf{H}}^{(\ell)}),
\end{align}
where $\mathbf{W}_{e}$ and $\mathbf{b}_{e}$ denote the linear embedding parameters.

\subsection{Quantformer-Style Numerical Embedding}
Unlike NLP models that transform discrete tokens into word embeddings, our input is already a structured numerical time series. Following Quantformer, we adopt a standard linear projection to embed the input sequence, instead of using token embedding and masking-based decoding. 

The embedding is defined as
\begin{equation}
\mathbf{H}^{(0)} = \mathbf{X}\mathbf{W}_{e} + \mathbf{b}_{e},
\end{equation}
where $\mathbf{W}_{e}\in\mathbb{R}^{C\times d}$ and $\mathbf{b}_{e}\in\mathbb{R}^{d}$.

\subsection{Temporal Self-Attention Encoder}
For each encoder block, multi-head self-attention is used to capture long-range temporal dependencies:
\begin{align}
\mathbf{Q}_{h} &= \mathbf{H}\mathbf{W}_{h}^{Q}, \quad
\mathbf{K}_{h} = \mathbf{H}\mathbf{W}_{h}^{K}, \quad
\mathbf{V}_{h} = \mathbf{H}\mathbf{W}_{h}^{V}, \\
\mathrm{Attn}_{h}(\mathbf{H}) &= \mathrm{Softmax}\left(
\frac{\mathbf{Q}_{h}\mathbf{K}_{h}^{\top}}{\sqrt{d_{h}}}
\right)\mathbf{V}_{h}.
\end{align}
The outputs of all heads are concatenated and linearly projected:
\begin{equation}
\mathrm{MHSA}(\mathbf{H}) =
\mathrm{Concat}(\mathrm{Attn}_{1},\dots,\mathrm{Attn}_{M})\mathbf{W}^{O}.
\end{equation}

\subsection{Lee Oscillator Activation Dynamics}
To enhance local nonlinear sensitivity, we incorporate the Lee Oscillator~\cite{lee2004transient, lee2019chaotic,lee2020quantum} dynamics into the activation stage. The oscillator evolves through a discrete-time excitatory--inhibitory process:
\begin{align}
E(t+1) &= f\!\left(a_1L(t)+a_2E(t)-a_3I(t)+a_4S(t)-\xi_E\right), \\
I(t+1) &= f\!\left(b_1L(t)-b_2E(t)-b_3I(t)+b_4S(t)-\xi_I\right), \\
\Omega(t+1) &= f(S(t)), \\
L(t) &= [E(t)-I(t)]\exp(-kS(t)^2)+\Omega(t),
\end{align}
where $f(\cdot)$ is implemented using a bounded nonlinear response, and $S(t)$ is the external stimulus induced by the pre-activation value. This design follows the discrete Lee Oscillator formulation used for chaotic temporal modeling.

\begin{table}[H]
\centering
\caption{Parameter settings of the \textbf{eight Lee Oscillator types} used in \textbf{QFCQT}.}
\label{tab:lee_oscillator_params}
\renewcommand{\arraystretch}{1.05}
\setlength{\tabcolsep}{2.8pt}
\resizebox{0.85\columnwidth}{!}{%
\begin{tabular}{c|ccccccccccccc}
\hline
\textbf{Type} & \boldmath$a_1$ & \boldmath$a_2$ & \boldmath$a_3$ & \boldmath$a_4$ & \boldmath$b_1$ & \boldmath$b_2$ & \boldmath$b_3$ & \boldmath$b_4$ & \boldmath$\xi_E$ & \boldmath$\xi_I$ & \boldmath$\mu$ & \boldmath$e$ & \boldmath$k$ \\
\hline
1 & 0.0  & 5.0  & 5.0  & 1.0  & 0.0  & -1.0 & 1.0  & 0.0  & 0.0 & 0.0 & 5.0 & 0.001 & 500 \\
2 & 0.5  & 0.55 & 0.55 & -0.5 & 0.5  & -0.55 & -0.55 & -0.5 & 0.0 & 0.0 & 1.0 & 0.001 & 50 \\
3 & 0.5  & 0.6  & 0.55 & 0.5  & -0.5 & -0.6  & -0.55 & 0.5  & 0.0 & 0.0 & 1.0 & 0.001 & 50 \\
4 & -0.5 & 0.55 & 0.55 & -0.5 & -0.5 & -0.55 & -0.55 & 0.5  & 0.0 & 0.0 & 1.0 & 0.001 & 50 \\
5 & -0.9 & 0.9  & 0.9  & -0.9 & 0.9  & -0.9  & -0.9  & 0.9  & 0.0 & 0.0 & 1.0 & 0.001 & 50 \\
6 & -0.9 & 0.9  & 0.9  & -0.9 & 0.9  & -0.9  & -0.9  & 0.9  & 0.0 & 0.0 & 1.0 & 0.001 & 300 \\
7 & -5.0 & 5.0  & 5.0  & -5.0 & 1.0  & -1.0  & -1.0  & 1.0  & 0.0 & 0.0 & 1.0 & 0.001 & 50 \\
8 & -5.0 & 5.0  & 5.0  & -5.0 & 1.0  & -1.0  & -1.0  & 1.0  & 0.0 & 0.0 & 1.0 & 0.001 & 300 \\
\hline
\end{tabular}
}
\end{table}

\begin{figure}[t]
    \centering
    \includegraphics[width=0.8\columnwidth]{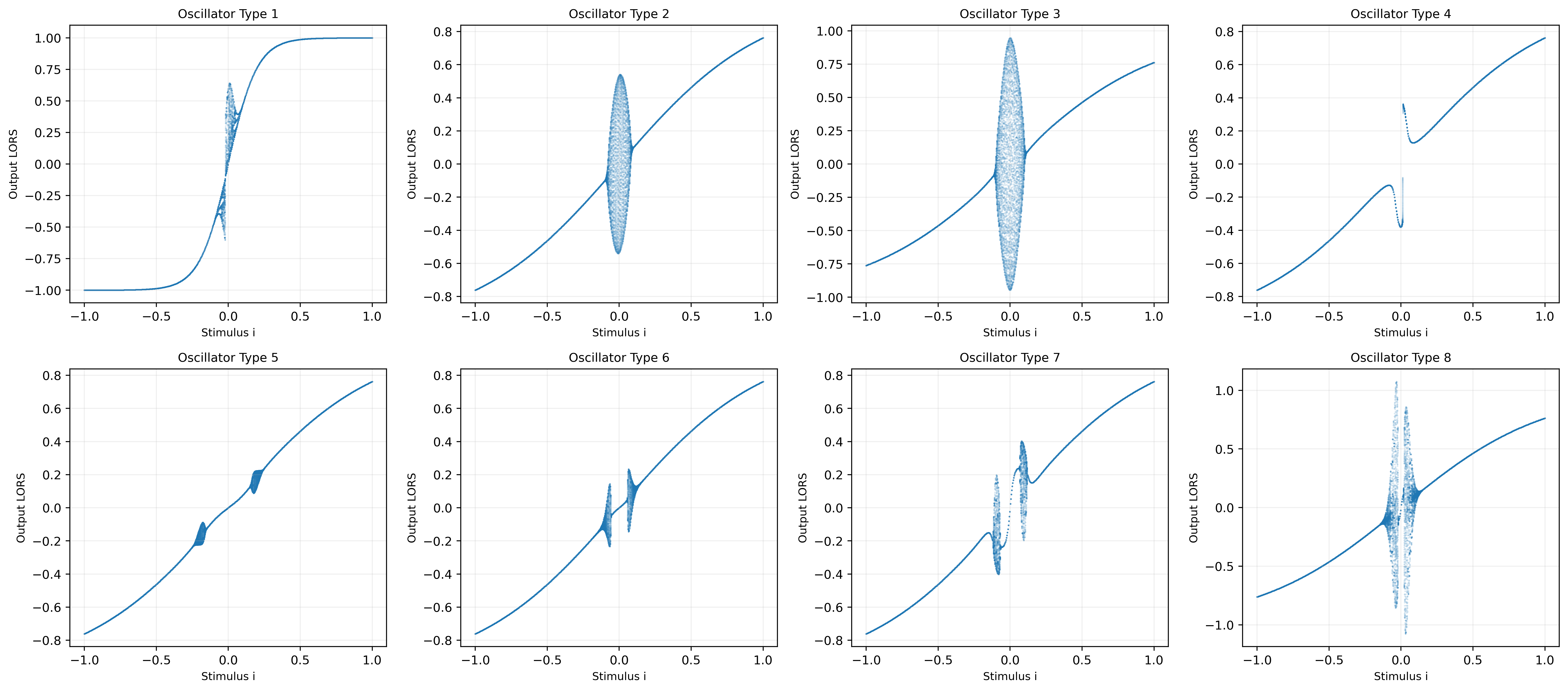}
\caption{Bifurcation visualizations of the eight Lee Oscillator types under the QFCQT parameter settings. Each subplot shows the dynamic output trajectories of one oscillator type with respect to the input stimulus, illustrating the diverse nonlinear and bifurcation behaviors captured by the proposed framework.}
\label{fig:oscillator_trajectories}
\end{figure}

The parameter settings in Table~\ref{tab:lee_oscillator_params} produce diverse nonlinear dynamics. Fig.~\ref{fig:oscillator_trajectories} shows that the eight oscillator types exhibit distinct bifurcation patterns, which enrich the response space of QFCQT for non-stationary time-series modeling.

\subsection{Max-over-Time Pooling}
A direct use of the full oscillator trajectory would dramatically increase dimensionality and computational cost. To obtain a practical scalar activation, we follow the COTN idea of compressing the internal temporal evolution through Max-over-Time pooling:
\begin{equation}
f_{k}^{\mathrm{MoT}}(x) = \max_{1 \le t \le N} L_{k}(t; x),
\end{equation}
where $k$ indexes the oscillator family and $N$ is the number of internal evolution steps.

\subsection{Soft Superposition of Eight Oscillator Families}
Instead of fixing a single oscillator type, we introduce a learnable soft combination over eight parameterized Lee oscillator families:
\begin{align}
\pi_k &= \frac{\exp(c_k)}{\sum_{j=1}^{8}\exp(c_j)}, \\
f_{\mathrm{chaos}}(x) &= \sum_{k=1}^{8}\pi_k f_{k}^{\mathrm{MoT}}(x).
\end{align}
This mechanism allows the model to adaptively choose suitable chaotic response patterns for different data regimes.

\subsection{Smooth-Chaotic Gated Fusion}
Although chaotic activation improves sensitivity, an overly aggressive nonlinear response may destabilize optimization. Following the gating philosophy of COTN, we fuse the chaotic activation with a smooth activation:
\begin{align}
f_{\mathrm{smooth}}(x) &= \mathrm{GELU}(x), \\
g &= \sigma(\lambda), \\
f_{\mathrm{QFCQT}}(x) &= g\,f_{\mathrm{smooth}}(x) + (1-g)\,f_{\mathrm{chaos}}(x),
\end{align}
where $\lambda$ is a learnable scalar gate. This fusion preserves the training stability of GELU while enabling the model to respond to highly volatile local perturbations. 

\begin{figure}[t]
    \centering
    \includegraphics[width=0.7\columnwidth]{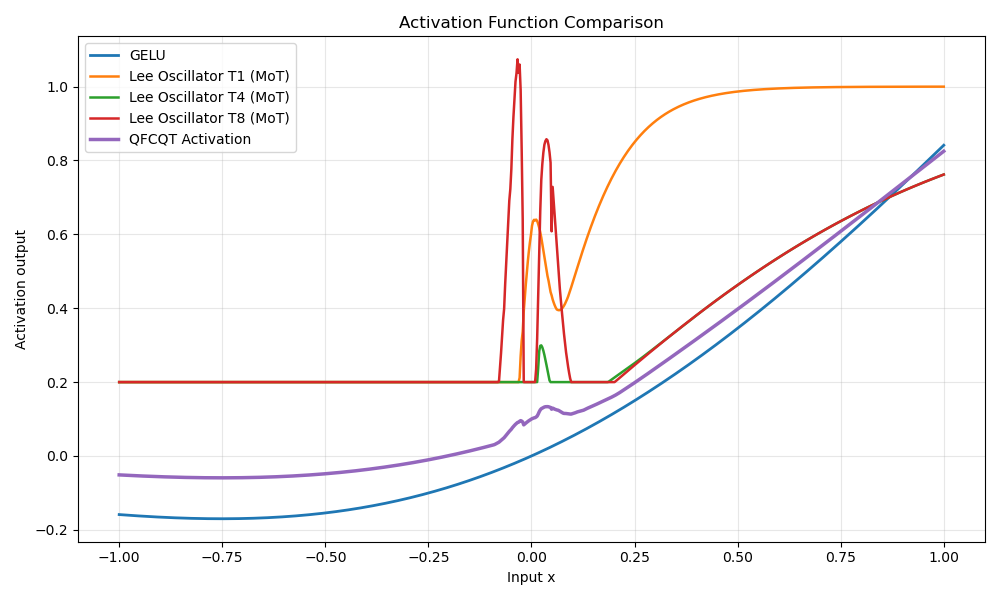}
    \caption{Activation response comparison of GELU, representative Lee-oscillator meta-activations, and the proposed QFCQT activation. QFCQT preserves smooth global behavior while introducing localized nonlinear sensitivity.}
    \label{fig:activation_comparison}
\end{figure}

\subsection{Chaotically Gated Feed-Forward Block}
To match the implementation, we use a channel expansion/projection structure with $1\times1$ convolution:
\begin{align}
\mathbf{U} &= \mathrm{Conv1D}^{\uparrow}_{1\times1}(\widetilde{\mathbf{H}}^{\top}), \\
\mathbf{V} &= f_{\mathrm{QFCQT}}(\mathbf{U}), \\
\mathbf{Z} &= \mathrm{Conv1D}^{\downarrow}_{1\times1}(\mathbf{V}), \\
\mathbf{H}^{(\ell)} &= \mathrm{LayerNorm}(\widetilde{\mathbf{H}} + \mathrm{Dropout}(\mathbf{Z}^{\top})).
\end{align}

\subsection{Optimization Objective}
For forecasting, we optimize the mean squared error:
\begin{equation}
\mathcal{L}_{\mathrm{MSE}} =
\frac{1}{HD}\sum_{t=1}^{H}\sum_{d=1}^{D}
(\widehat{Y}_{t,d}-Y_{t,d})^2,
\end{equation}
while MAE is used as an additional evaluation metric:
\begin{equation}
\mathcal{L}_{\mathrm{MAE}} =
\frac{1}{HD}\sum_{t=1}^{H}\sum_{d=1}^{D}
|\widehat{Y}_{t,d}-Y_{t,d}|.
\end{equation}

\section{Experimental Configuration}

\subsection{Datasets Description \& Preprocessing}

To evaluate the proposed model across different domains, we use two representative time-series datasets. The first is the \textbf{ETT (Electricity Transformer Temperature)} dataset released by the State Grid Corporation of China, which reflects real-world power system dynamics through oil temperature and load-related variables. In this work, we use two hourly subsets, namely \textbf{ETTh$_1$} and \textbf{ETTh$_2$}, each containing 8,640 records. Both subsets exhibit strong volatility and nonlinear temporal patterns, making them standard benchmarks for challenging forecasting tasks. In particular, transformer faults, overload conditions, and abrupt load variations may induce severe fluctuations, thus imposing higher demands on model robustness.

For the financial domain, we use a \textbf{high-frequency A-share stock dataset} from a private data vendor, containing more than 17,000 one-minute records over several months. Each record includes OHLC prices and trading volume. The dataset captures key intraday market characteristics, such as volatility clustering, liquidity variation, and rapid short-term fluctuations, making it suitable for evaluating forecasting performance under extreme financial conditions.

The preprocessing pipeline includes data cleaning and feature construction. Timestamp continuity is maintained by forward-filling short missing intervals and removing longer discontinuities. Duplicate records are discarded, and abnormal points are filtered using domain-specific rules together with Z-score thresholds. Based on the OHLCV data, we further construct standard temporal features, including log-returns, moving averages, and volatility-related indicators.
\subsection{Baselines}
We compare against representative baselines including Informer, LogTrans, LSTMa, HAT, COTN, and TimesNet. These baselines cover efficient Transformer forecasting, recurrent sequence modeling, and oscillator-enhanced Transformer modeling.

\subsection{Metrics}
We report Mean Squared Error (MSE) and Mean Absolute Error (MAE), where lower values indicate better forecasting performance. To highlight the advantage of QFCQT, we additionally report relative improvements over HAT and COTN, and TimesNet.

\section{Experimental Results}

\subsection{Overall Performance}
Table~\ref{tab:main_results_simple} and~\ref{tab:main_results} reports the comprehensive forecasting results on ETTh$_1$, ETTh$_2$, and A-share Stock Index datasets.

\begin{table}[t]
\centering
\caption{Simplified performance comparison on \textbf{ETT} and \textbf{A-share Stock Index}. Best results are highlighted in \textbf{bold}.}
\label{tab:main_results_simple}
\renewcommand{\arraystretch}{1.08}
\setlength{\tabcolsep}{3.2pt}
\footnotesize

\resizebox{0.85\columnwidth}{!}{%
\begin{tabular}{c c|cc|cc|cc}
\hline
\textbf{Dataset} & \textbf{Hor.}
& \multicolumn{2}{c|}{\textbf{HAT}}
& \multicolumn{2}{c|}{\textbf{COTN}}
& \multicolumn{2}{c}{\textbf{QFCQT}} \\
\cline{3-8}
& & MSE & MAE & MSE & MAE & MSE & MAE \\
\hline

\multirow{4}{*}{ETTh$_1$}
& 24  & 0.576 & 0.558 & 0.548 & 0.522 & \textcolor{lightgreen}{\textbf{0.484}} & \textcolor{lightgreen}{\textbf{0.492}} \\
& 48  & 0.689 & 0.614 & 0.649 & \textcolor{lightgreen}{\textbf{0.566}} & \textcolor{lightgreen}{\textbf{0.619}} & 0.584 \\
& 168 & 0.930 & 0.749 & 0.865 & 0.683 & \textcolor{lightgreen}{\textbf{0.619}} & \textcolor{lightgreen}{\textbf{0.584}} \\
& 336 & 1.244 & 0.857 & 1.146 & \textcolor{lightgreen}{\textbf{0.771}} & \textcolor{lightgreen}{\textbf{0.971}} & 0.795 \\
\hline

\multirow{4}{*}{ETTh$_2$}
& 24  & 0.719 & 0.662 & 0.686 & 0.623 & \textcolor{lightgreen}{\textbf{0.403}} & \textcolor{lightgreen}{\textbf{0.487}} \\
& 48  & 1.445 & 0.984 & 1.364 & 0.915 & \textcolor{lightgreen}{\textbf{1.163}} & \textcolor{lightgreen}{\textbf{0.823}} \\
& 168 & 3.476 & 1.490 & 3.250 & 1.351 & \textcolor{lightgreen}{\textbf{2.023}} & \textcolor{lightgreen}{\textbf{1.035}} \\
& 336 & 2.718 & 1.327 & \textcolor{lightgreen}{\textbf{2.514}} & 1.198 & 2.542 & \textcolor{lightgreen}{\textbf{1.182}} \\
\hline

\multirow{3}{*}{A-share}
& 24  & 3.328 & 1.371 & 3.174 & 1.298 & \textcolor{lightgreen}{\textbf{3.073}} & \textcolor{lightgreen}{\textbf{1.207}} \\
& 96  & 3.584 & 1.523 & 3.394 & 1.476 & \textcolor{lightgreen}{\textbf{3.196}} & \textcolor{lightgreen}{\textbf{1.337}} \\
& 336 & 3.694 & 1.601 & 3.453 & 1.476 & \textcolor{lightgreen}{\textbf{3.209}} & \textcolor{lightgreen}{\textbf{1.344}} \\
\hline
\end{tabular}
}
\end{table}

\begin{table*}[hbt!]
\centering
\caption{Model Comprehensiveness Evaluation for \textbf{ETT} and \textbf{A-share Stock Index}}
\label{tab:main_results}
\renewcommand{\arraystretch}{1.0}      % 从 1.1 改为 1.0
\setlength{\tabcolsep}{3pt}            % 或 2.8pt
\scriptsize                            % 替代 \footnotesize（如果当前没设置）
\resizebox{\textwidth}{!}{
\begin{tabular}{c c|cc|cc|cc|cc|cc|cc|cc}
\hline
\textbf{Dataset} & \textbf{Horizon} 
& \multicolumn{2}{c|}{\textbf{Informer}}
& \multicolumn{2}{c|}{\textbf{LogTrans}}
& \multicolumn{2}{c|}{\textbf{LSTMa}}
& \multicolumn{2}{c|}{\textbf{HAT}}
& \multicolumn{2}{c|}{\textbf{COTN}}
& \multicolumn{2}{c|}{\textbf{TimesNet}}
& \multicolumn{2}{c}{\textbf{\textcolor{ieeeblue}{QFCQT (Ours)}}} \\
\cline{3-16}
& & MSE & MAE & MSE & MAE & MSE & MAE 
& MSE & MAE & MSE & MAE 
& MSE & MAE
& MSE & MAE \\
\hline

\multirow{4}{*}{ETTh$_1$}
& 24 
& 0.577 & 0.549 & 0.686 & 0.604 & 0.650 & 0.624 
& 0.576 & 0.558 & 0.548 & 0.522
& 0.592 & 0.585
& \textbf{\textcolor{ieeeblue}{0.484}} & \textbf{\textcolor{ieeeblue}{0.492}} \\

& 48 
& 0.685 & 0.625 & 0.766 & 0.757 & 0.702 & 0.675 
& 0.689 & 0.614 & 0.649 & \textbf{0.566}
& 0.745 & 0.672
& \textbf{\textcolor{ieeeblue}{0.619}} & 0.584 \\

& 168 
& 0.931 & 0.752 & 1.002 & 0.846 & 1.212 & 0.867 
& 0.930 & 0.749 & 0.865 & 0.683
& 0.845 & 0.758
& \textbf{\textcolor{ieeeblue}{0.619}} & \textbf{\textcolor{ieeeblue}{0.584}} \\

& 336 
& 1.128 & 0.873 & 1.362 & 0.952 & 1.424 & 0.994 
& 1.244 & 0.857 & 1.146 & \textbf{0.771}
& 1.205 & 0.915
& \textbf{\textcolor{ieeeblue}{0.971}} & 0.795 \\
\hline

\multirow{4}{*}{ETTh$_2$}
& 24 
& 0.720 & 0.665 & 0.828 & 0.750 & 1.143 & 0.813 
& 0.719 & 0.662 & 0.686 & 0.623
& 0.528 & 0.615
& \textbf{\textcolor{ieeeblue}{0.403}} & \textbf{\textcolor{ieeeblue}{0.487}} \\

& 48 
& 1.457 & 1.001 & 1.806 & 1.034 & 1.671 & 1.211 
& 1.445 & 0.984 & 1.364 & 0.915
& 1.425 & 0.976
& \textbf{\textcolor{ieeeblue}{1.163}} & \textbf{\textcolor{ieeeblue}{0.823}} \\

& 168 
& 3.489 & 1.515 & 4.070 & 1.681 & 4.117 & 1.675 
& 3.476 & 1.490 & 3.250 & 1.351
& 2.815 & 1.355
& \textbf{\textcolor{ieeeblue}{2.023}} & \textbf{\textcolor{ieeeblue}{1.035}} \\

& 336 
& 2.723 & 1.340 & 3.875 & 1.763 & 3.434 & 1.549
& 2.718 & 1.327 & \textbf{2.514} & 1.198
& 2.955 & 1.385
& 2.542 & \textbf{\textcolor{ieeeblue}{1.182}} \\
\hline

\multirow{3}{*}{A-share Stock Index}
& 24 
& 3.335 & 1.381 & 3.435 & 1.477  
& 3.546 & 1.570 & 3.328 & 1.371 
& 3.174 & 1.298
& 3.295 & 1.298
& \textbf{\textcolor{ieeeblue}{3.073}} & \textbf{\textcolor{ieeeblue}{1.207}} \\

& 96 
& 3.608 & 1.567 & 3.727 & 1.671 
& 4.038 & 1.835 & 3.584 & 1.523 
& 3.394 & 1.476
& 3.415 & 1.425
& \textbf{\textcolor{ieeeblue}{3.196}} & \textbf{\textcolor{ieeeblue}{1.337}} \\

& 336 
& 3.702 & 1.620 & 3.754 & 1.670  
& 4.657 & 2.105 & 3.694 & 1.601 
& 3.453 & 1.476
& 3.475 & 1.418
& \textbf{\textcolor{ieeeblue}{3.209}} & \textbf{\textcolor{ieeeblue}{1.344}} \\
\hline
\end{tabular}
}
\end{table*}

\begin{table}[t]
\centering
\caption{Relative Improvement of \textbf{QFCQT} over Baseline Models}
\label{tab:improvement_results}
\renewcommand{\arraystretch}{1.05}
\setlength{\tabcolsep}{2.2pt}
\scriptsize
\resizebox{0.8\columnwidth}{!}{
\begin{tabular}{c c|cc|cc|cc}
\hline
\textbf{Dataset} & \textbf{Hor.}
& \multicolumn{2}{c|}{\textbf{Inc. HAT}}
& \multicolumn{2}{c|}{\textbf{Inc. COTN}}
& \multicolumn{2}{c}{\textbf{Inc. TimesNet}} \\
\cline{3-8}
& & MSE & MAE & MSE & MAE & MSE & MAE \\
\hline

\multirow{4}{*}{ETTh$_1$}
& 24  
& \textcolor{lightgreen}{16.0$\uparrow$} & \textcolor{lightgreen}{11.8$\uparrow$}
& \textcolor{lightgreen}{11.7$\uparrow$} & \textcolor{llg}{5.7$\uparrow$}
& \textcolor{lightgreen}{18.2$\uparrow$} & \textcolor{lightgreen}{15.9$\uparrow$} \\

& 48  
& \textcolor{lightgreen}{10.1$\uparrow$} & \textcolor{llg}{4.9$\uparrow$}
& \textcolor{llg}{4.6$\uparrow$} & \textcolor{gray}{$\sim$ (-3.2)}
& \textcolor{lightgreen}{16.9$\uparrow$} & \textcolor{lightgreen}{13.1$\uparrow$} \\

& 168 
& \textcolor{darkgreen}{33.4$\uparrow$} & \textcolor{darkgreen}{22.0$\uparrow$}
& \textcolor{darkgreen}{28.4$\uparrow$} & \textcolor{lightgreen}{14.5$\uparrow$}
& \textcolor{darkgreen}{26.7$\uparrow$} & \textcolor{darkgreen}{23.0$\uparrow$} \\

& 336 
& \textcolor{darkgreen}{21.9$\uparrow$} & \textcolor{llg}{7.2$\uparrow$}
& \textcolor{lightgreen}{15.2$\uparrow$} & \textcolor{gray}{$\sim$ (-3.1)}
& \textcolor{lightgreen}{19.4$\uparrow$} & \textcolor{lightgreen}{13.1$\uparrow$} \\
\hline

\multirow{4}{*}{ETTh$_2$}
& 24  
& \textcolor{darkgreen}{43.9$\uparrow$} & \textcolor{darkgreen}{26.5$\uparrow$}
& \textcolor{darkgreen}{41.3$\uparrow$} & \textcolor{darkgreen}{21.8$\uparrow$}
& \textcolor{darkgreen}{23.7$\uparrow$} & \textcolor{darkgreen}{20.8$\uparrow$} \\

& 48  
& \textcolor{lightgreen}{19.5$\uparrow$} & \textcolor{lightgreen}{16.3$\uparrow$}
& \textcolor{lightgreen}{14.8$\uparrow$} & \textcolor{lightgreen}{10.0$\uparrow$}
& \textcolor{lightgreen}{18.4$\uparrow$} & \textcolor{lightgreen}{15.7$\uparrow$} \\

& 168 
& \textcolor{darkgreen}{41.79$\uparrow$} & \textcolor{darkgreen}{30.57$\uparrow$}
& \textcolor{darkgreen}{37.74$\uparrow$} & \textcolor{darkgreen}{23.43$\uparrow$}
& \textcolor{darkgreen}{28.1$\uparrow$} & \textcolor{darkgreen}{23.6$\uparrow$} \\

& 336 
& \textcolor{llg}{6.48$\uparrow$} & \textcolor{llg}{10.93$\uparrow$}
& \textcolor{gray}{$\sim$ (-1.1)} & \textcolor{llg}{1.34$\uparrow$}
& \textcolor{lightgreen}{14.0$\uparrow$} & \textcolor{lightgreen}{14.7$\uparrow$} \\
\hline

\multirow{3}{*}{A-share}
& 24  
& \textcolor{llg}{7.7$\uparrow$} & \textcolor{lightgreen}{12.0$\uparrow$}
& \textcolor{llg}{3.2$\uparrow$} & \textcolor{llg}{7.0$\uparrow$}
& \textcolor{llg}{6.7$\uparrow$} & \textcolor{llg}{7.0$\uparrow$} \\

& 96  
& \textcolor{lightgreen}{10.8$\uparrow$} & \textcolor{lightgreen}{12.2$\uparrow$}
& \textcolor{llg}{5.8$\uparrow$} & \textcolor{llg}{9.4$\uparrow$}
& \textcolor{llg}{6.4$\uparrow$} & \textcolor{llg}{6.2$\uparrow$} \\

& 336 
& \textcolor{lightgreen}{13.1$\uparrow$} & \textcolor{lightgreen}{16.0$\uparrow$}
& \textcolor{llg}{7.1$\uparrow$} & \textcolor{llg}{8.9$\uparrow$}
& \textcolor{llg}{7.7$\uparrow$} & \textcolor{llg}{5.2$\uparrow$} \\
\hline
\end{tabular}
}
\end{table}

As shown in Table~\ref{tab:main_results_simple} and~\ref{tab:main_results}, QFCQT achieves the best or competitive performance in most settings across ETTh$_1$, ETTh$_2$, and A-share Stock Index datasets. The improvement is most significant on ETTh$_2$, especially at horizons 24 and 168, indicating that the proposed chaotically gated activation is more beneficial under highly volatile and nonlinear dynamics. On the A-share dataset, the gains are more moderate but consistent, suggesting that the framework can transfer from electricity benchmarks to financial forecasting scenarios.

\subsection{Comparison Against HAT and COTN}
The comparison with HAT and COTN is particularly informative. HAT is a strong Transformer-style baseline, while COTN already incorporates chaotic activation ideas. Compared with HAT, QFCQT shows that a Quantformer-style numerical backbone combined with chaos-aware activation yields systematic improvements across both energy and financial datasets. Compared with COTN, QFCQT demonstrates that simply introducing a chaotic unit is not sufficient; the way it is integrated into the numerical backbone also matters.

Across ETTh$_1$ and ETTh$_2$, QFCQT generally achieves larger gains in MSE than in MAE, indicating that it is especially effective in reducing large prediction deviations and improving robustness under difficult segments. The largest relative improvements occur precisely on the volatile ETTh$_2$ settings, which supports the central motivation of our method: chaotic gating is particularly beneficial when the data contains abrupt changes and nonlinear oscillatory patterns. On the A-share dataset, the gains are more moderate but consistent, suggesting that the framework can transfer from electricity benchmarks to financial forecasting scenarios.

\subsection{Ablation Analysis of QFCQT on the ETT Benchmark}

To better understand the contribution of each core component in QFCQT, we conduct a comprehensive ablation study on the ETT benchmark for Horizon is 24. Specifically, three key modules are progressively removed or simplified: 

1) the quantum-gated chaotic activation module, including 
\textbf{QuantumSuperposition\_LORS} and \textbf{VectorizedLeeOscillator}; 

2) the fractal-modulated attention module, i.e., 
\textbf{FractalModulatedAttention}; and 

3) the original forecasting head, which is replaced with a simpler linear regression head. In this way, the ablated variant preserves the overall training protocol and backbone configuration while removing the major non-linear and structure-enhancing designs of the full model.

The purpose of this ablation setting is to examine whether the performance gains of QFCQT mainly come from its coupled chaotic activation mechanism, its attention design, or the prediction head. By comparing the complete model against the simplified variant, we can quantitatively assess the effectiveness of these components under different forecasting horizons. The corresponding results on ETTh$_1$ and ETTh$_2$ are reported in Table~\ref{table:ablation_results}.

As shown in Table~\ref{table:ablation_results}, removing these components leads to consistent performance degradation on both ETTh$_1$ and ETTh$_2$, which confirms that each of them contributes positively to the forecasting capability of QFCQT. Nevertheless, even after the above simplifications, the ablated model still remains competitive on short-horizon forecasting, and in some cases continues to outperform representative baselines such as HAT and COTN.

\begin{table}[t]
\centering
\caption{Comparison between the full \textbf{QFCQT} model and its ablated variant on ETTh$_1$ and ETTh$_2$.}
\label{table:ablation_results}
\setlength{\tabcolsep}{6pt}
\renewcommand{\arraystretch}{1.1}
\begin{tabular}{|c|c|c|c|}
\hline
\textbf{Model} & \textbf{Test MSE} & \textbf{Test MAE} & \textbf{Effect of Ablation} \\
\hline
\textbf{Full (ETTh$_1$)} 
& \cellcolor{green!20}\textbf{0.484} 
& \cellcolor{green!20}\textbf{0.492} 
& \cellcolor{green!20}\textbf{Reference} \\
\hline
\textbf{Ablation (ETTh$_1$)} 
& \cellcolor{red!20}0.487 
& \cellcolor{red!20}0.494 
& \cellcolor{red!20}Performance Drop \\
\hline
\textbf{Full (ETTh$_2$)} 
& \cellcolor{green!20}\textbf{0.403} 
& \cellcolor{green!20}\textbf{0.487} 
& \cellcolor{green!20}\textbf{Reference} \\
\hline
\textbf{Ablation (ETTh$_2$)} 
& \cellcolor{red!20}0.478 
& \cellcolor{red!20}0.532 
& \cellcolor{red!20}Performance Drop \\
\hline
\end{tabular}
\end{table}

Table~\ref {table:ablation_results} shows that removing the proposed nonlinear components leads to consistent degradation on both ETTh$_1$ and ETTh$_2$. The performance drop is small on ETTh$_1$ but more evident on ETTh$_2$, suggesting that the oscillator-based gated design is particularly useful under stronger volatility. Even after ablation, the simplified variant remains competitive with HAT and COTN at horizon 24. This confirms that both the Quantformer-style numerical backbone and the chaotically gated activation contribute to the final performance.

\subsection{Why the Proposed Method Works}
The performance gains of QFCQT can be attributed to three complementary factors.

First, the Quantformer-style numerical embedding avoids unnecessary NLP-style assumptions and is better aligned with ordered quantitative observations. This provides a cleaner and more direct representation for numerical sequence forecasting. 

Second, the Lee Oscillator activation introduces a richer nonlinear response landscape than conventional pointwise activations. Through internal temporal evolution and Max-over-Time pooling, the model can transform each scalar pre-activation into a dynamic meta-activation that is more sensitive to sharp local variations. 

Third, the smooth-chaotic gated fusion stabilizes training. Instead of letting chaotic responses dominate all hidden transformations, the learnable gate adaptively balances the oscillator response and the smooth GELU component. This controlled fusion allows the model to benefit from enhanced responsiveness without sacrificing convergence behavior. 

\section{Discussion}
These results suggest that feed-forward activation design is an important but often underexplored factor in Transformer forecasting. QFCQT does not claim a formal quantum or fractal theory; rather, the term “quantum-fractal-inspired” refers to a computational analogy based on soft oscillator-family superposition and multi-scale nonlinear response organization. By constraining chaotic responses through Max-over-Time pooling and smooth-chaotic gated fusion, the model improves sensitivity to abrupt local changes while maintaining stable optimization.

\section{Conclusion}
In this paper, we proposed QFCQT, a chaotically gated Quantformer framework for time-series forecasting under complex volatile dynamics. Unlike conventional Transformer-based forecasting models relying on fixed smooth activations, QFCQT integrates a Lee-oscillator-based dynamic activation, Max-over-Time pooling, and a smooth-chaotic gated fusion strategy into a Quantformer-style encoder, enabling effective modeling of long-range dependencies and nonlinear temporal dynamics.

The proposed framework was evaluated on ETTh$_1$, ETTh$_2$, and A-share Stock Index benchmarks. Experimental results showed that QFCQT consistently outperformed representative baselines, including Informer, LogTrans, LSTMa, HAT, COTN, and TimesNet, across multiple forecasting horizons, with particularly notable improvements under highly volatile settings.

Overall, the results demonstrate that the feed-forward nonlinear transformation plays a crucial role in Transformer forecasting. By introducing adaptive chaotic dynamics, QFCQT provides an effective solution for non-stationary time-series prediction, and future work will explore its extension to broader sequence modeling tasks.
\bibliographystyle{IEEEtran}
\bibliography{ref}
\nocite{*}

\end{document}